\documentclass[lettersize,journal]{IEEEtran}
\usepackage{amsmath,amsfonts}
\usepackage{algorithmic}
\usepackage{algorithm}
\usepackage{array}
\usepackage[caption=false,font=normalsize,labelfont=sf,textfont=sf]{subfig}
\usepackage{textcomp}
\usepackage{stfloats}
\usepackage{url}
\usepackage{verbatim}
\usepackage{graphicx}
\usepackage{cite}
\usepackage{booktabs}
\usepackage{multirow}
\usepackage{pifont}
\usepackage{xcolor} 
\usepackage[table]{xcolor}
\usepackage{balance}    
\usepackage[most]{tcolorbox}

\newtcolorbox{promptbox}[1][]{
  colback=gray!5!white,   
  colframe=gray!75!black, 
  title=\textbf{System Prompt}, 
  fonttitle=\bfseries,
  sharp corners,
  boxrule=0.5pt,
  left=5pt, right=5pt, top=5pt, bottom=5pt,
  #1
}
\definecolor{LightBlue}{rgb}{0.9, 0.95, 1.0}

\AtBeginDocument{%
  \providecommand\BibTeX{{%
    \normalfont B\kern-0.5em{\scshape i\kern-0.25em b}\kern-0.8em\TeX}}}

\begin{document}

\title{Eval-Actions: Fine-Grained Execution Quality Evaluation for Robotic Manipulation}

\author{Mengyuan Liu, Juyi Sheng$^{\ast}$, Peiming Li, Ziyi Wang, Tianming Xu, Tiantian Xu, Hong Liu$^{\ast}$
        
\thanks{Anonymous Project Website at https://eval-actions.github.io/.}
}

\markboth{Regular Paper}%
{Shell \MakeLowercase{\textit{et al.}}: A Sample Article Using IEEEtran.cls for IEEE Journals}

\maketitle

\begin{abstract}
Although Vision--Action (VA) and Vision--Language--Action (VLA) policies have advanced robotic manipulation, their evaluation remains dominated by binary success rates, which obscure process-level differences among executions that complete the same task. We introduce Eval-Actions, a diagnostic evaluation methodology and real-robot benchmark for fine-grained execution-quality assessment of learned manipulation policies. Eval-Actions combines criteria-based Expert Grading (EG), Rank-Guided (RG) labels that align measurable motion indicators with expert rankings, and Chain-of-Thought-style (CoT) annotations that explain observable quality differences. The benchmark contains 13K+ teleoperated and policy-generated real-robot episodes covering 150+ tasks and approximately 52 hours of recordings with RGB-D videos, robot-state trajectories, task descriptions, and success/failure labels. Its densely annotated subset provides EG/RG/CoT supervision for training and evaluation. We further provide AutoEval, a reference multimodal evaluator that predicts quality scores, task outcomes, and diagnostic explanations from RGB temporal evidence and compact kinematic summaries. On the annotated Eval-Actions test split, AutoEval-S achieves Spearman rank correlations (SRCCs) of 0.81 and 0.84 under EG and RG, with success detection accuracies of 90.6\% and 91.0\%; AutoEval-P reaches 0.70 SRCC under CoT. Analyses of expert consistency, physical-metric baselines, modality ablations, structured generalization, and offline policy ranking show that Eval-Actions provides standardized, interpretable diagnostic signals complementary to success-rate evaluation.
\end{abstract}


\section{Introduction}
\label{intro}

\IEEEPARstart{R}{ecent} advances in imitation learning and data-driven robot learning have accelerated the development of learned manipulation policies. A recent survey summarizes progress in robot learning from multimodal demonstrations~\cite{tro0}. Meanwhile, low-cost data-collection systems such as ALOHA~\cite{aloha} and UMI~\cite{umi}, together with public datasets such as RLBench~\cite{rlbench} and Open X-Embodiment~\cite{openx}, have lowered the barrier to collecting real-robot demonstrations and training manipulation policies. These advances have further driven Vision--Action (VA) and Vision--Language--Action (VLA) policies, including ACT~\cite{aloha}, Diffusion Policy~\cite{dp1,dp2}, HDP~\cite{tro1}, MP1~\cite{sheng2025mp1}, RT-1~\cite{rt1}, RT-2~\cite{rt2}, RT-X~\cite{openx}, OpenVLA~\cite{openvla}, and $\pi_0$~\cite{pi0}. However, evaluation protocols for robot learning policies have not progressed at the same pace. Most manipulation studies still rely primarily on binary task success, i.e., whether the robot reaches a target state such as grasping, inserting, stacking, or placing an object. Such outcome-only metrics provide limited information about how the outcome is achieved, especially with respect to motion smoothness, contact behavior, repeated attempts, visible collision-related events, and execution time.

\begin{figure*}[t]
    \centering
    \includegraphics[width=1.0\linewidth]{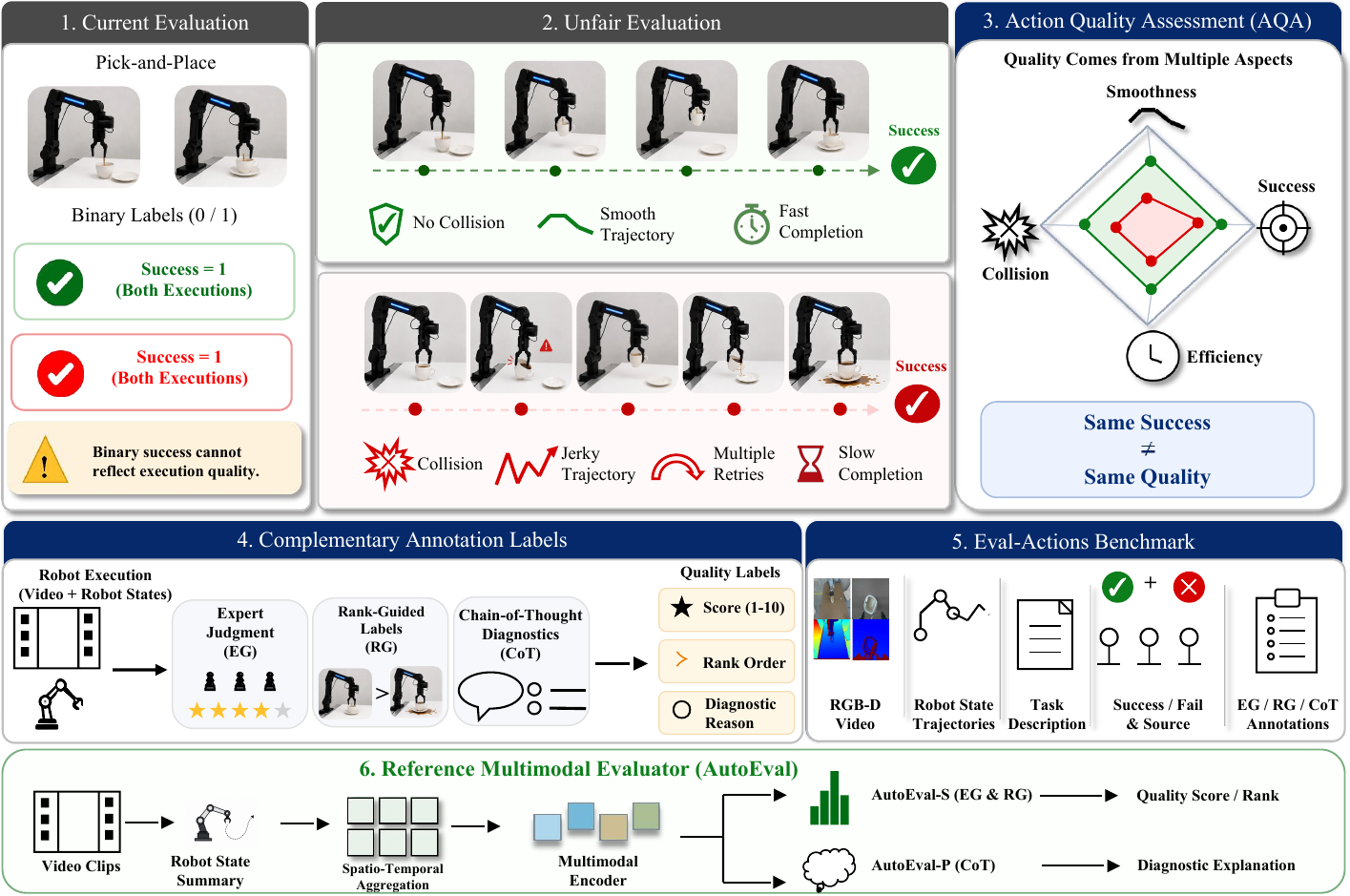}

\caption{Motivation and overview of the Eval-Actions diagnostic evaluation framework.
(1) Conventional manipulation evaluation often reduces each execution to a binary success/failure label, which cannot distinguish process quality among successful executions.
(2) Two executions may both complete the same task while differing in motion smoothness, visible collision-related events, repeated retries, and execution efficiency.
(3) We formulate robotic manipulation evaluation as an Action Quality Assessment (AQA) problem, where execution quality is assessed from task completion, motion smoothness, execution efficiency, and visible collision-related events.
(4) Eval-Actions provides three complementary annotations: criteria-based Expert Grading (EG), Rank-Guided (RG) labels, and CoT-style diagnostic explanations.
(5) Eval-Actions stores RGB-D videos, robot-state trajectories, task descriptions, success/failure labels, auxiliary trajectory-source metadata, and EG/RG/CoT annotations.
(6) AutoEval is provided as a reference multimodal evaluator: AutoEval-S predicts quality scores under the EG/RG protocols, while AutoEval-P generates CoT-style diagnostic explanations.}
    \label{challenge}
\end{figure*}

\begin{table*}[t]
\centering

    \caption{Comparison of representative robotic manipulation datasets. Existing datasets target policy training, focusing on trajectory quantity and diversity. Eval-Actions is designed for diagnostic evaluation and includes failure cases, mixed trajectory sources, fine-grained quality scores, and CoT-style explanatory annotations based on RGB-D videos and robot trajectories.}
\label{tab:dataset_comparison_en}

\resizebox{\textwidth}{!}{%
\begin{tabular}{ll c c c c c c c c} 
\toprule
\multirow{2}{*}{\textbf{Dataset}} & \multirow{2}{*}{\textbf{Year}} & \multirow{2}{*}{\textbf{Focus}} & \textbf{Raw} & \multicolumn{3}{c}{\textbf{Diagnostic Annotations}} & \multirow{2}{*}{\textbf{Arm}} & \multirow{2}{*}{\textbf{Collection}} & \multirow{2}{*}{\textbf{Modalities}} \\
\cmidrule(lr){5-7}
 & & & \textbf{Traj.} & \textbf{Failures} & \textbf{Scoring} & \textbf{CoT} & & & \\
\midrule
BridgeData V2 \cite{walke2023bridgedata} & 2023 & Training & 60k & \ding{55} & \ding{55} & \ding{55} & Single & Human+Script & RGB-D, Text \\
DobbE \cite{bringing} & 2023 & Training & 5.6k & \ding{55} & \ding{55} & \ding{55} & Single & Tool-based & RGB-D, Text \\
Open X-Embodiment \cite{openx} & 2024 & Training & \textbf{1M+} & \ding{55} & \ding{55} & \ding{55} & Mixed & Aggregation & RGB-D, PC, Text \\
RH20T \cite{RH20T} & 2024 & Training & 13k & \ding{55} & \ding{55} & \ding{55} & Single & Teleop. & RGB-D, Force, Audio \\
DROID \cite{khazatsky2024droid} & 2024 & Training & 76k & \ding{55} & \ding{55} & \ding{55} & Single & Teleop. & RGB-D, Text \\
RoboMIND \cite{wu2024robomind} & 2025 & Training & 107k & 1.6k & \ding{55} & \ding{55} & Mixed & Teleop. & RGB-D, Text \\
\midrule
\rowcolor{gray!15} 
\textbf{Eval-Actions (Ours)} & -- & \textbf{Evaluation} & 13k & \textbf{2.8k} & \textbf{\ding{51}} & \textbf{\ding{51}} & Mixed & \textbf{Hybrid (Teleop. \& Policy)} & \textbf{RGB-D, Text} \\
\bottomrule 
\\
\multicolumn{9}{l}{
    \parbox{\linewidth}{
        \footnotesize \itshape 
        $^{\ast}$Note: RoboMIND's failure data primarily consists of re-grasping events rather than genuine terminal failures. Note that 'Scoring' and 'CoT' refer to the diagnostic annotations proposed in this work rather than inherent properties of the original raw datasets.
    }
}

\end{tabular}%
}
\end{table*}

\begin{figure*}[t]
    \centering
    \includegraphics[width=1.0\linewidth]{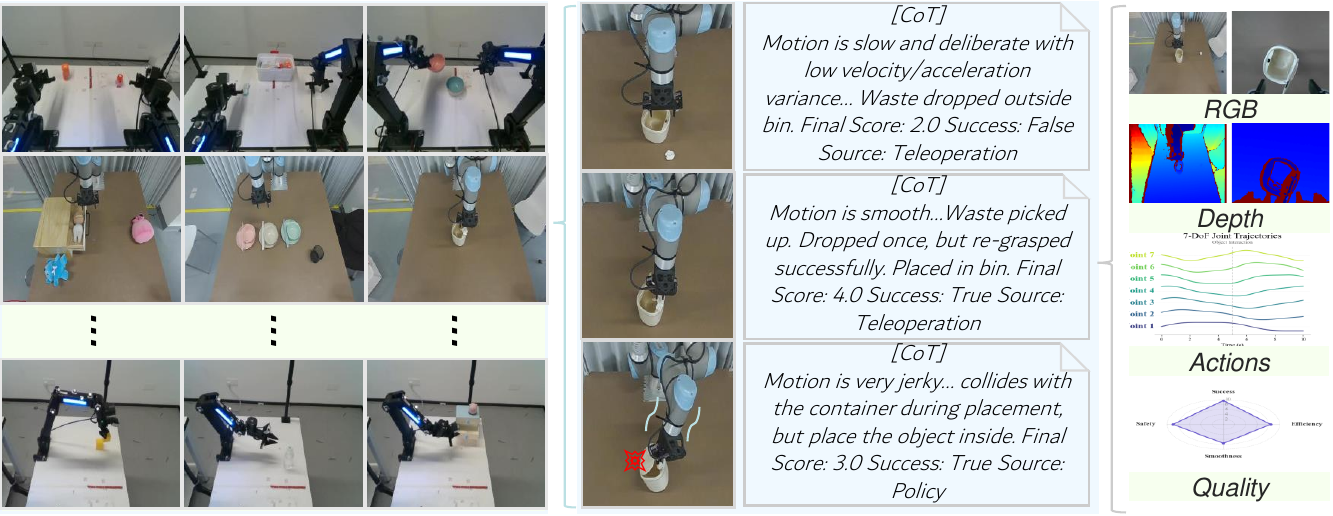}
    \caption{\textbf{Overview of the Eval-Actions Benchmark.} 
    The figure illustrates the dataset structure:
    \textbf{(Left) Task Diversity:} Representative snapshots from 150+ scenarios, covering single-arm interactions (e.g., “Throw away trash”) and bimanual coordination tasks (e.g., “Tidy medicine box”).
    \textbf{(Middle) Detailed Case Study:} A specific instantiation of the “Throw away trash” task. Each task includes diverse demonstration data ranging from high-quality successes to failures, enabling comparison between smooth teleoperation and jerky policy executions.
    \textbf{(Right) Data Composition:} The stack enumerates the dense multimodal signals within each episode, including raw sensory data (RGB, Depth), precise kinematic records (7/14-DoF joint trajectories), and the Fine-Grained Quality Radar Chart, which quantifies four core dimensions (Task Completion, Smoothness, Collision-related Events, Efficiency) for diagnostic evaluation.}
    \label{dataset_overview}
\end{figure*}

The limitation of binary success becomes particularly pronounced in real-robot deployment and policy diagnosis. As illustrated in Fig.~\ref{challenge}-(2), two executions may complete the same manipulation objective while exhibiting substantially different execution quality. For example, one policy may place a cup at the target location only after multiple retries, abrupt motions, and unstable contacts, whereas another may complete the same task smoothly and efficiently. Although both executions receive the same success label, their deployment reliability, contact-related execution characteristics, and diagnostic value are different. To capture this missing dimension, we formulate robotic manipulation evaluation as a fine-grained Action Quality Assessment (AQA) problem: given an execution video and robot-state trajectory, the evaluator estimates execution-process quality while also determining the task outcome. Unlike conventional human-centric AQA~\cite{aqa2,aqa3}, robotic manipulation quality is more closely tied to task completion, motion stability, visible collision-related events, execution efficiency, and temporal consistency. Robotic AQA, therefore, requires the joint use of visual temporal evidence and robot trajectory information, while explicitly distinguishing criteria-based expert judgments from measurable kinematic indicators.

Based on this observation, our central contribution is not merely to collect a trajectory dataset, but to define and instantiate a triadic evaluation methodology for robotic AQA. We decompose execution-quality assessment into three complementary supervision views: Expert Grading (EG), Rank-Guided (RG) labels, and Chain-of-Thought-style (CoT) diagnostic annotations. EG captures criteria-based expert judgment under a unified scoring rubric; RG calibrates measurable physical indicators with expert rankings to obtain metric-grounded quality labels; and CoT annotations record observable diagnostic evidence that explains the assigned quality assessment. This EG/RG/CoT protocol connects expert preference, explicit physical indicators, and interpretable diagnostic rationales in a single evaluation setting.

A key obstacle to establishing such a triadic evaluation methodology is the absence of a unified protocol and a corresponding real-robot benchmark instantiation for execution-quality analysis. Existing robotic manipulation datasets are primarily designed for policy training. As shown in Table~\ref{tab:dataset_comparison_en}, they typically emphasize trajectory scale, task diversity, or aggregation across robot embodiments, but rarely provide failed executions, successful samples with low execution quality, fine-grained quality scores, preference rankings, and explanatory diagnostic annotations under a unified evaluation protocol. To address this gap, we introduce Eval-Actions as an EG/RG/CoT-based evaluation methodology and instantiate it as a diagnostic real-robot benchmark. Eval-Actions contains RGB-D videos, robot action trajectories, task descriptions, success/failure labels, and trajectory-source metadata for controlled source-aware analysis of teleoperated and policy-generated executions. To support controlled training and evaluation, we further construct an annotated subset with sufficient coverage of failures and low-quality successes. This subset implements the proposed protocol by providing EG, RG, and CoT annotations, enabling systematic analysis of execution quality beyond binary task success. The trajectory-source labels are used only for source-aware analysis within the benchmark and should not be interpreted as provenance or authenticity verification.

To operationalize the proposed EG/RG/CoT evaluation methodology, we introduce AutoEval as a reference multimodal evaluator for robotic manipulation executions. AutoEval predicts quality scores, task outcomes, and diagnostic explanations defined by the Eval-Actions protocol from RGB temporal evidence and compact robot-kinematic summaries. AutoEval-S performs structured prediction under the EG and RG protocols, using Spatio-Temporal Aggregation to preserve short-term motion cues within a fixed VLM input budget, while AutoEval-P targets CoT diagnostic explanation generation and improves consistency between explanations and structured predictions. On the Eval-Actions test split, AutoEval-S achieves Spearman rank correlation coefficients (SRCCs) of 0.81 and 0.84 under the EG and RG protocols, respectively, while AutoEval-P obtains an SRCC of 0.70 under the CoT protocol. We further validate Eval-Actions and AutoEval through expert-annotation consistency analysis, explicit physical-metric baselines, temporal aggregation and modality ablations, structured generalization, and offline policy-level ranking of representative VA/VLA policies. The main contributions of this paper are summarized as follows:

\begin{itemize}
    \item We formulate the evaluation of learned VA/VLA manipulation policies as a fine-grained robotic Action Quality Assessment problem and introduce an EG/RG/CoT triadic evaluation methodology for assessing execution-process quality beyond binary task success.

    \item We instantiate this methodology in Eval-Actions, a diagnostic real-robot benchmark that provides RGB-D recordings, robot-state trajectories, task descriptions, success/failure labels, trajectory-source metadata for source-aware analysis, and EG/RG/CoT annotations for teleoperated and policy-generated executions.

    \item We provide AutoEval as a reference multimodal evaluator for operationalizing the proposed protocol, and validate its diagnostic utility for robot learning policy evaluation through in-distribution evaluation, structured generalization, physical-metric comparisons, modality ablations, and offline policy-level ranking.
\end{itemize}

\section{Related Work}
\label{sec:Related}


\subsection{Robot Learning Datasets and Evaluation Benchmarks}
Large-scale datasets are fundamental for data-driven robotic manipulation, providing vast demonstrations through teleoperation~\cite{ebert2021bridge, walke2023bridgedata}, distributed collection~\cite{bringing, khazatsky2024droid}, cross-embodiment aggregation~\cite{openx}, and multitask settings~\cite{RH20T, wu2024robomind}. However, these resources are primarily designed for policy training or outcome-level evaluation. They rarely provide failed executions, fine-grained quality scores, or diagnostic annotations. Eval-Actions directly addresses this gap by treating each execution as an evaluation instance, providing EG, RG, and CoT annotations alongside multimodal trajectories to diagnose process-level quality—such as motion smoothness and execution efficiency—beyond binary success.


\subsection{VA/VLA Policies and Their Evaluation}
Manipulation policies have rapidly evolved from specialized primitives~\cite{grasp1, tro2, tro_GRASP} to end-to-end Vision--Action (VA)~\cite{aloha, dp1, dp2, dp3, sheng2025mp1} and Vision--Language--Action (VLA) foundation models~\cite{rlbench, peract, rvt, rt1, rt2, openvla, octo, pi0, pi05}. Despite these architectural advances, evaluation protocols remain largely outcome-centric. Binary success rates indicate whether a goal is reached but obscure how the robot reaches it. Consequently, two policies with identical success rates may exhibit vastly different contact behaviors or retries. Orthogonal to policy design, Eval-Actions and AutoEval provide a unified diagnostic protocol to explicitly evaluate these process-level discrepancies.


\subsection{Action Quality Assessment for Robotic Manipulation}
Action Quality Assessment (AQA) estimates execution quality rather than mere categorization. While widely applied in computer vision for sports~\cite{aqa2, aqa3, pan2025basket} and surgical skill evaluation~\cite{liu2021towards, SEDSkill}, robotic AQA differs significantly. Instead of aesthetics or posture, robotic execution quality depends strictly on task completion, motion stability, execution efficiency, and collision-related events, requiring the joint analysis of visual and kinematic evidence. This work adapts AQA for robot learning by introducing EG, RG, and CoT, formulating a unified framework for process-level policy evaluation.

\section{The Eval-Actions Dataset}
\label{sec:dataset}

Eval-Actions is designed as a diagnostic real-robot benchmark for fine-grained, video-based execution-quality evaluation of learned VA/VLA manipulation policies. Unlike training-centric datasets that primarily provide demonstration trajectories for policy learning, Eval-Actions treats each real robot execution as an evaluation instance and supports systematic diagnostic analysis of the execution process. Specifically, Eval-Actions provides evaluation signals for task outcome, process-level quality scoring, and diagnostic explanation, together with trajectory-source metadata for controlled analysis of teleoperated and policy-generated executions. The trajectory-source metadata is used as auxiliary analysis information rather than as a primary definition of execution quality.

\subsection{Dataset Construction}

As shown in Fig~\ref{dataset_overview}, Eval-Actions covers a wide range of single-arm and dual-arm manipulation scenarios, such as object grasping, bowl stacking, table cleaning, medicine-box organization, and plate handover. The dataset contains approximately 52 hours of real-robot manipulation recordings, including more than 13k episodes across over 150 tasks. The data are collected on heterogeneous robot platforms, including ARX R5 and UR5 manipulators, covering both single-arm execution and bimanual coordination.

Eval-Actions contains two complementary trajectory sources: human teleoperation from 20 operators with diverse backgrounds, and executions generated by representative VA/VLA policies. This hybrid collection strategy is used to cover a broader spectrum of execution behaviors, including smooth successes, successful executions with low quality, partial failures, and complete failures. The dataset also provides trajectory-source metadata to support controlled source-aware analysis; this metadata is not used to define the EG or RG execution-quality labels.

Compared with datasets that primarily emphasize successful demonstrations, Eval-Actions intentionally includes failure cases and successful executions with low quality. These episodes are important for evaluating action quality beyond binary success, because a successful trajectory may still contain retries, jitter, inefficient motion, or visible collision-related events. In total, Eval-Actions contains 2.8k failed executions, which exposes evaluators to a wider range of execution-quality levels rather than only high-quality demonstrations.

Each episode stores synchronized multimodal information, including RGB-D recordings from wrist-mounted and/or third-person cameras, robot joint states and end-effector states, action trajectories, textual task descriptions, binary success labels, and trajectory-source metadata. For annotated episodes, diagnostic annotations include EG, RG, and CoT explanations, which are detailed in Sec.~\ref{sec:annotation}. EG provides expert quality scores under a unified rubric; RG calibrates measurable motion indicators using expert rankings collected under the same evaluation rubric; and CoT explanations describe how different evaluation criteria affect the final score.

To support reproducible and controlled evaluation, we define a densely annotated subset, Eval-Actions Small (EAS; detailed in Table~\ref{tab:dataset_statistics}), for training and evaluating AutoEval. EAS includes failed executions and successful executions with low quality, and each episode has complete EG, RG, CoT, success, and trajectory-source metadata. Unless otherwise specified, all AutoEval results are reported on EAS. The data are split into disjoint training, validation, and test partitions, and the RG calibration parameters are estimated only on the training split before being fixed and applied to validation and test episodes. In addition to the standard in-distribution split, structured generalization splits are constructed for joint unseen-task-and-arm, unseen-task-only, and unseen-arm-only evaluation.
\begin{figure}[t]
    \centering
    \includegraphics[width=1.0\linewidth]{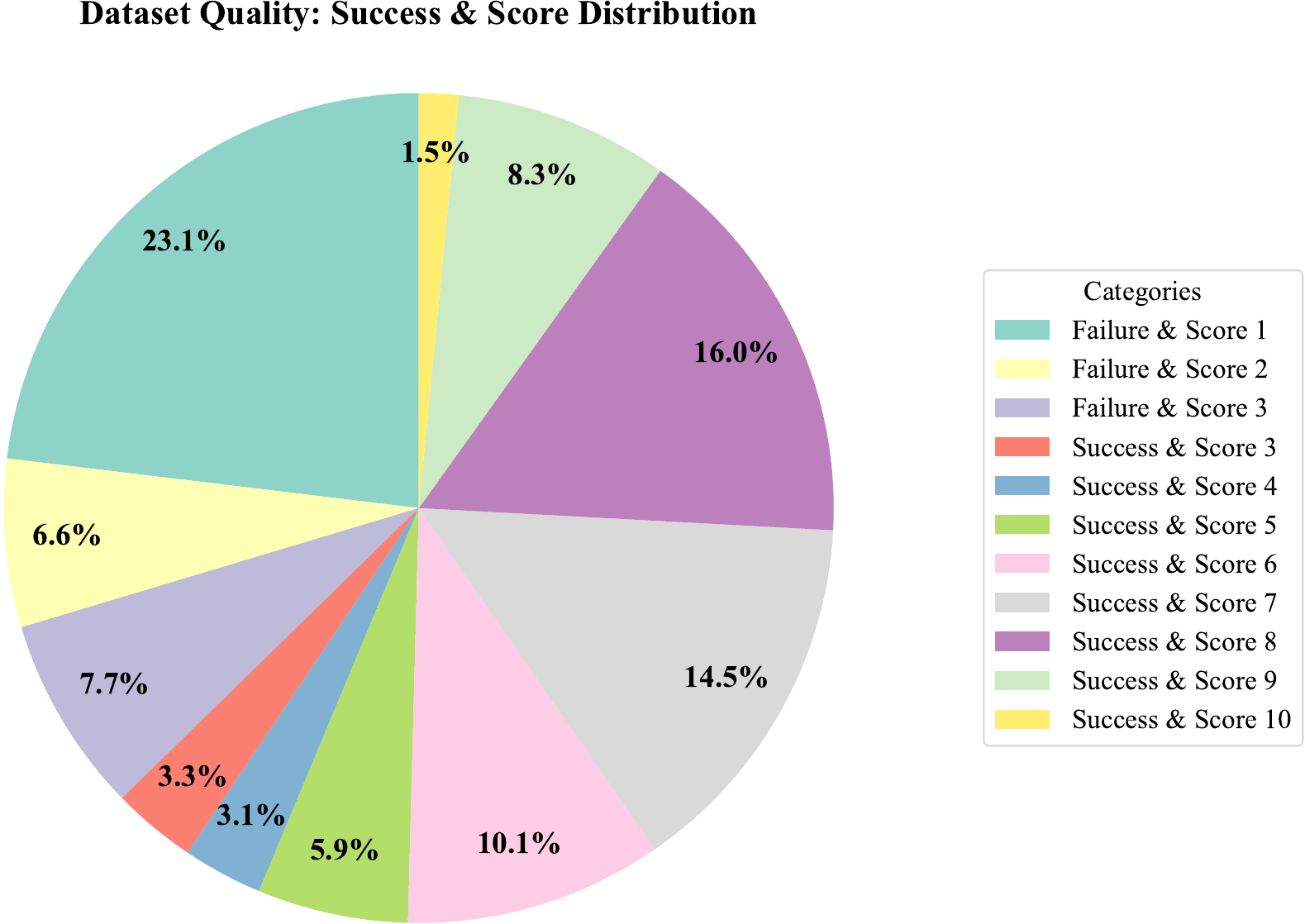}

\caption{Distribution of Expert Grading (EG) scores in the Eval-Actions Small (EAS) subset. The chart shows the proportions of failed executions and successful executions at different quality levels over the 1--10 score range.}
    \label{statics}
\end{figure}

\subsection{Data Annotation and Definition of Fine-Grained Action Quality}
\label{sec:annotation}

We define \textit{Fine-Grained Action Quality} as a criteria-based assessment of the execution process for each annotated robotic manipulation episode. The annotation protocol is built upon four predefined criteria: \textbf{task completion}, \textbf{motion smoothness}, \textbf{visible collision-related events}, and \textbf{execution efficiency}. Task completion is recorded as a binary success/failure label. Motion smoothness is characterized by robot-state statistics, including the variation of joint angular velocity and joint angular acceleration. Visible collision-related events refer to unintended contacts, collisions, object drops, or abnormal interactions that can be identified from videos or robot logs when available. This criterion is used as evidence for execution quality rather than as a complete safety certification. Execution efficiency is measured by task completion time and normalized within the same task and initialization region to reduce the influence of different starting distances.

Based on the above criteria, Eval-Actions provides three complementary annotation signals: EG, RG, and CoT diagnostic annotations.

\textbf{EG.}
EG is obtained through criteria-based expert scoring. For each annotated episode, 10 expert annotators inspect the RGB video, task description, and task outcome, and assign an integer quality score from 1 to 10 according to unified criteria, including task success, motion smoothness, visible collision-related events, and completion efficiency. Policy identity and trajectory-source metadata are not used for EG scoring. Scores from 1--3 correspond to low-quality executions, scores from 4--7 correspond to acceptable or moderate-quality executions, and scores from 8--10 correspond to high-quality executions. The stored EG label is the average score across annotators. The complete scoring rubric and ambiguity-handling rules are provided in \textbf{Appendix-III}; EG annotation reliability is analyzed in Sec.~\ref{sec:human_human_srcc}.

\textbf{RG.}
RG calibrates measurable motion indicators using expert rankings, producing a metric-grounded quality label. Experts first rank small batches of execution videos according to overall action quality using the same criteria as EG. We then define a rule-based score based on task completion, normalized smoothness indicators, visible collision-related penalties, and normalized execution time. The weights of these terms are optimized with a Genetic Algorithm (GA) on the training split, so that the ranking induced by the rule-based score is as close as possible to the expert ranking. The learned weights are then fixed and used to generate RG labels for the validation and test splits, avoiding information leakage during label construction. Therefore, RG is treated as a metric-grounded diagnostic label rather than an absolute physical ground truth.

\textbf{CoT.}
CoT annotations provide textual explanations of how the predefined evaluation criteria affect the final quality assessment. Each explanation describes observable evidence, such as whether the task is completed, whether retries occur, whether the motion is abrupt, whether the object is dropped, whether visible collisions or abnormal contacts occur, and whether the execution is inefficient. These explanations are paired with the quality score and success label, enabling the evaluator to output interpretable diagnostic results in addition to scalar scores. Trajectory-source metadata is serialized as auxiliary information for source-aware analysis, but it does not define the EG/RG quality scores. CoT annotations are used as auxiliary explanation targets, while EG and RG remain the primary protocols for quantitative action quality assessment.

\subsection{Expert Annotation Reliability}
\label{sec:human_human_srcc}

To assess the reliability of EG annotations, we analyze both ranking consistency and numerical rating reliability. Each annotated robot execution episode is independently scored by 10 expert annotators on a 1--10 scale. Annotators are provided with the RGB video, task description, and task outcome; policy identity and trajectory-source metadata are not used for EG scoring. The scoring criteria include task completion, motion smoothness, visible collision-related events, and execution efficiency. Visible collision-related events are used only as observable evidence for execution quality rather than as a complete safety certification. For ambiguous cases, annotators first consider task completion and then adjust the score according to smoothness, visible collision-related events, and efficiency. RG labels and CoT annotations are constructed separately and are not used to determine the EG score.

We compute human-human Spearman rank correlation coefficient (SRCC) using a leave-one-expert-out protocol, and use the intraclass correlation coefficient (ICC) to measure numerical rating reliability. This protocol does not designate any single expert as the absolute ground truth; instead, it measures how well an individual expert agrees with the consensus of the remaining expert panel. The detailed computation is provided in \textbf{Appendix IV}. As shown in Table~\ref{tab:inter_expert_consistency}, the EG scores achieve a leave-one-expert-out human-human SRCC of $0.91 \pm 0.06$, indicating strong ranking agreement between individual experts and the remaining expert consensus. The ICC(2,1) value is $0.88$, and ICC(2,K) reaches $0.99$, suggesting that individual expert ratings are already reliable and that averaging over multiple experts further improves the stability of EG scores. Therefore, we use the aggregated expert score as a reference annotation for training and evaluating AutoEval, while avoiding treating any single expert as the absolute ground truth.

\begin{table}[t]
\centering
\caption{Inter-expert consistency of EG annotations.}
\label{tab:inter_expert_consistency}
\small
\setlength{\tabcolsep}{3pt}
\renewcommand{\arraystretch}{1.08}
\begin{tabular*}{\columnwidth}{@{\extracolsep{\fill}}l c c c c@{}}
\toprule
\textbf{Annotation} & \textbf{K} & \textbf{LOO SRCC} 
& \textbf{ICC(2,1)} & \textbf{ICC(2,K)} \\
\midrule
EG score & 10 & $0.91 \pm 0.06$ & $0.88$ & $0.99$ \\
\bottomrule
\end{tabular*}
\vspace{0.5mm}
\parbox{\columnwidth}{
\footnotesize
\textit{Note:} All metrics are computed on the 1--10 EG scores. LOO SRCC denotes leave-one-expert-out human-human Spearman rank correlation, reported as mean $\pm$ standard deviation across experts. ICC(2,1) measures the reliability of a single expert rating, while ICC(2,K) measures the reliability of the averaged score from $K$ experts.
}
\vspace{-4pt}
\end{table}

\subsection{Dataset Statistics and Splits}
\label{sec:dataset_statistics}

The full Eval-Actions benchmark contains 13K+ episodes, 150+ tasks, and approximately 52 hours of recordings. EAS is the annotation-complete subset, containing 6K+ episodes, 50+ tasks, and approximately 12 hours of recordings, with a 37.4\% failure ratio. EG scores in EAS cover a broad 1--10 range, allowing the benchmark to distinguish high-quality successes from low-quality successes involving retries, jitter, visible contact, or inefficient motion.

\begin{table}[t]
\centering
\caption{Statistics of Eval-Actions and EAS.}
\label{tab:dataset_statistics}
\footnotesize
\setlength{\tabcolsep}{2.0pt}
\renewcommand{\arraystretch}{1.05}
\begin{tabular*}{\columnwidth}{@{\extracolsep{\fill}}lccccc@{}}
\toprule
\textbf{Dataset} & \textbf{Eps.} & \textbf{Tasks} & \textbf{Hrs.} 
& \textbf{Fail.} & \textbf{Split} \\
\midrule
Eval-Actions & 13K+ & 150+ & 52 & 2.8K & -- \\
EAS & 6K+ & 50+ & 12 & 37.4\% & 80/10/10 \\
\bottomrule
\end{tabular*}
\vspace{0.5mm}
\begin{minipage}{0.98\columnwidth}
\scriptsize
\textit{Note:} Fail. denotes failed-episode count for Eval-Actions and failure ratio for EAS. EAS is used for AutoEval training and evaluation unless otherwise specified.
\end{minipage}
\vspace{-6pt}
\end{table}

\subsection{Label-Level Agreement with Expert Grading}
\label{sec:direct_physical_label}

To justify the need for expert-guided calibration, we examine whether action-quality labels can be directly constructed from physical indicators without using expert ranking preferences. For each episode in the held-out test split, we compute a candidate label \(S_{\mathrm{cand}}\) and measure its rank agreement with the expert-consensus EG score \(S_{\mathrm{EG}}\):
\begin{equation}
\rho_{\mathrm{cand}\text{-}\mathrm{EG}}
=
\mathrm{SRCC}
\left(
\{S_{\mathrm{cand},i}\}_{i\in\mathcal{D}_{\mathrm{test}}},
\{S_{\mathrm{EG},i}\}_{i\in\mathcal{D}_{\mathrm{test}}}
\right).
\end{equation}
This is a label-level agreement test rather than a model prediction experiment. For the binary success label, we directly use the task outcome. For the direct physical score, we use the same normalized physical indicators as RG but with fixed manually assigned weights and penalties, without expert-ranking calibration. For RG, the physical indicators are retained, but their weights and penalties are calibrated using expert rankings on the training split and then fixed for the held-out test split.

\begin{table}[t]
\centering
\caption{
Label-level agreement between candidate labels and expert grading.
The first three rows report rank agreement with the aggregated EG score on the held-out test split, while human-human agreement is reported using the leave-one-expert-out protocol.
}
\label{tab:direct_physical_label_agreement}
\small
\setlength{\tabcolsep}{4pt}
\renewcommand{\arraystretch}{1.08}
\begin{tabular*}{\columnwidth}{@{\extracolsep{\fill}}lcc@{}}
\toprule
\textbf{Candidate label} & \textbf{Human guidance} & \textbf{Rank agreement} $\uparrow$ \\
\midrule
Binary success label & No & 0.53 \\
Direct physical score & No & 0.78 \\
RG label (ours) & Expert ranking & 0.90 \\
Human-human agreement & Expert panel & $0.91 \pm 0.06$ \\
\bottomrule
\end{tabular*}
\vspace{-4pt}
\end{table}

As shown in Table~\ref{tab:direct_physical_label_agreement}, binary success achieves only 0.53 SRCC with EG, indicating that task outcome alone is too coarse for execution-quality assessment. Direct physical scoring improves the agreement to 0.78 SRCC, showing that physical indicators provide meaningful and reproducible proxy labels. However, it still falls below the human-human agreement of \(0.91 \pm 0.06\), suggesting that fixed physical weights and penalties cannot fully capture expert-perceived quality.

In contrast, the RG label reaches 0.90 SRCC, close to human-human agreement. This indicates that physical metrics are useful, but their relative importance should be calibrated by human quality preferences. Therefore, human-guided labels are introduced not to replace physical metrics, but to align measurable physical indicators with human-perceived execution quality.

\begin{figure*}[t]
    \centering
    \includegraphics[width=1.0\linewidth]{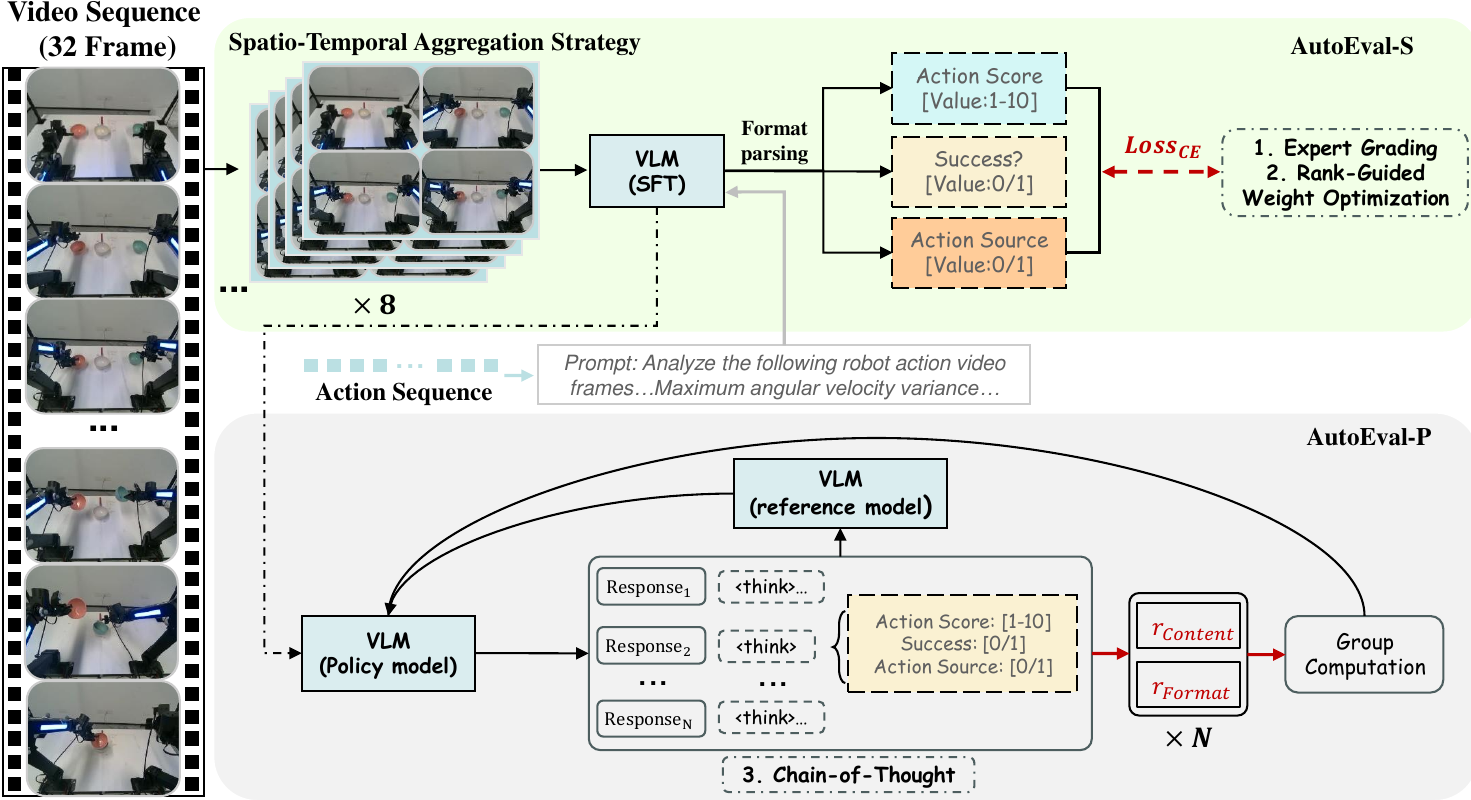}
    \caption{\textbf{Overview of the proposed AutoEval framework.}
    AutoEval takes a robot manipulation video sequence and a compact kinematic summary as inputs.
    \textbf{Top (AutoEval-S):} AutoEval-S is used for structured prediction under the EG and RG protocols. The Spatio-Temporal Aggregation strategy tiles neighboring frames into composite visual inputs, preserving short-term motion cues under a fixed VLM visual-input budget. The model outputs the action quality score, task success label, and trajectory source label, and is optimized by supervised fine-tuning with a cross-entropy loss.
    \textbf{Bottom (AutoEval-P):} AutoEval-P is used for CoT-style diagnostic explanation generation. This branch is initialized from the SFT model and optimized with Group Relative Policy Optimization (GRPO)~\cite{GPRO, guo2025deepseek}. Multiple candidate responses are evaluated using a hybrid reward composed of content accuracy ($r_{\mathrm{Content}}$) and format constraints ($r_{\mathrm{Format}}$), encouraging consistency between diagnostic explanations and structured predictions.}
    \label{autoeval}
\end{figure*}

\section{Method}

In this section, we present AutoEval, a VLM-based reference evaluator for diagnosing the execution quality of robotic manipulation episodes. AutoEval is trained with the supervision signals provided by Eval-Actions, including EG, RG, and CoT diagnostic annotations. Given RGB temporal evidence and compact robot-kinematic summaries, AutoEval predicts execution quality, task outcome, and diagnostic explanations. We first summarize Rank-Guided Weight Optimization, which constructs RG labels by aligning measurable motion indicators with expert rankings. We then introduce two evaluator variants: AutoEval-S for structured prediction under the EG/RG protocols, and AutoEval-P for CoT diagnostic explanation generation. Trajectory-source metadata is treated as an auxiliary output for controlled source-aware analysis, rather than as a definition of execution quality.

\subsection{\textit{Rank-Guided Weight Optimization}}
\label{subsec:optimization}

Determining the relative contribution of measurable indicators to perceived execution quality is non-trivial. To construct RG labels, we align a weighted physical score with expert rankings on the training split. Let $\boldsymbol{\theta}=\{w_1,\ldots,w_M,\lambda_{\mathrm{coll}},\lambda_{\mathrm{fail}}\}$ denote the tunable parameters, where $w_i$ are the weights of normalized quality indicators and $\lambda_{\mathrm{coll}}$, $\lambda_{\mathrm{fail}}$ are penalty factors for visible collision-related events and task failure. For each normalized indicator $s_i$, larger values correspond to better execution quality. When a violation is observed, the corresponding indicator is penalized before aggregation:
\begin{equation}
    S_{\mathrm{raw}}(\boldsymbol{\theta})
    =
    \frac{\sum_{i=1}^{M} w_i s'_i}{\sum_{i=1}^{M} w_i},
    \quad
    s'_i =
    \frac{s_i}{\eta_i},
\end{equation}
where $\eta_i \in \{1, \lambda_{\mathrm{coll}}, \lambda_{\mathrm{fail}}, 
\lambda_{\mathrm{coll}}\lambda_{\mathrm{fail}}\}$ is determined by whether collision-related or task-completion violations apply to the indicator.

We use a Genetic Algorithm (GA) to search for the parameters that minimize the discrepancy between the ranking induced by $S_{\mathrm{raw}}$ and the expert consensus ranking $R_{\mathrm{exp}}$:
\begin{equation}
    \boldsymbol{\theta}^{\ast}
    =
    \arg\min_{\boldsymbol{\theta}}
    \frac{1}{B}
    \sum_{b=1}^{B}
    \left|
    R_{\mathrm{exp}}^{(b)}
    -
    R_{\mathrm{raw}}^{(b)}(\boldsymbol{\theta})
    \right|,
\end{equation}
where $B$ is the number of trajectories used for calibration. Since this objective optimizes relative ordering, the resulting raw scores are aligned to the 1--10 EG scale using training-split statistics:
\begin{equation}
    S_{\mathrm{RG}}
    =
    \mu_{\mathrm{EG}}
    +
    \sigma_{\mathrm{EG}}
    \left(
    \frac{
    S_{\mathrm{raw}}(\boldsymbol{\theta}^{\ast})-\mu_{\mathrm{raw}}
    }{
    \sigma_{\mathrm{raw}}+\epsilon
    }
    \right).
\end{equation}
The optimized weights and affine alignment parameters are fixed and applied to validation and test episodes, avoiding information leakage during RG label construction. Detailed GA implementation settings are provided in \textbf{Appendix V}.

\subsection{AutoEval}

AutoEval is a unified VLM-based evaluator with two variants for the annotation protocols in Eval-Actions. AutoEval-S performs structured prediction under the EG and RG protocols, while AutoEval-P generates CoT diagnostic explanations. Both variants use RGB temporal evidence and a compact kinematic summary constructed from robot joint states.

\textbf{Task Formulation.}
For a manipulation episode with $T$ time steps, the visual input is represented by a sequence of sampled frames $\mathcal{F}=\{f_i\}_{i=1}^{N}$, where $N \leq T$. The robot kinematic trajectory is represented by a joint state matrix $\mathbf{Q}\in\mathbb{R}^{T\times J}$, where $J\in\{7,14\}$ denotes the degrees of freedom. After synchronizing the joint states to a uniform time grid, the angular velocity $\mathbf{v}_t$ and angular acceleration $\boldsymbol{\alpha}_t$ are computed using discrete differences:
\begin{equation}
\mathbf{v}_t = \mathbf{q}_{t} - \mathbf{q}_{t-1}, \quad
\boldsymbol{\alpha}_t = \mathbf{v}_{t} - \mathbf{v}_{t-1},
\label{eq:kinematics}
\end{equation}
where $\mathbf{q}_t\in\mathbb{R}^{J}$ denotes the joint configuration at time step $t$.

To provide explicit motion cues, we compute temporal variance vectors for joint velocity and acceleration, denoted as $\mathbf{s}_v,\mathbf{s}_\alpha\in\mathbb{R}^{J}$. We summarize them using worst-case smoothness indicators, defined as the maximum variance across joints, together with the mean absolute velocity:
\begin{equation}
\begin{split}
\mathcal{U}_v &= \max_{j \in \{1,\dots,J\}} \mathbf{s}_{v}[j], \quad
\mathcal{U}_\alpha = \max_{j \in \{1,\dots,J\}} \mathbf{s}_{\alpha}[j], \\
\mu_v &= \frac{1}{(T-1)J} \sum_{t=2}^{T} \sum_{j=1}^{J} |v_{t,j}|.
\end{split}
\label{eq:uniformity}
\end{equation}

These metrics are serialized into a structured textual descriptor $I_{\mathrm{phys}}$, which provides a compact kinematic summary for motion smoothness and intensity. The multimodal evaluator $\Phi_\theta$ takes the visual frames and the kinematic descriptor as input, and predicts the action quality score $\hat{S}$, task outcome $\hat{O}$, and an auxiliary trajectory-source tag $\hat{C}$:
\begin{equation}
(\hat{S}, \hat{O}, \hat{C})
=
\Phi_\theta\left(
\mathcal{F}, I_{\mathrm{phys}}(\mathcal{U}_v,\mathcal{U}_\alpha,\mu_v)
\right).
\label{eq:model_inference}
\end{equation}
Here, $\hat{S}$ corresponds to the quality score under the current annotation protocol, $\hat{O}$ denotes the predicted success/failure label, and $\hat{C}$ denotes the predicted trajectory source.

\textbf{AutoEval-S.}
A straightforward way to improve temporal reasoning is to increase the number of sampled frames, but this also increases the number of visual tokens and memory cost. To preserve local motion cues within a fixed VLM input budget, AutoEval-S uses a \textit{Spatio-Temporal Aggregation} strategy, as illustrated in the top of Fig.~\ref{autoeval}. Instead of representing each sampled time step by a single frame, we tile $k$ neighboring frames with the target frame into a composite image and resize it to the standard encoder resolution. This produces an aggregated visual sequence $\mathcal{F}'=\{f'_i\}_{i=1}^{N}$, allowing each visual input to encode short-term temporal changes.

For optimization, action quality assessment, task success detection, and trajectory source classification are formulated as a unified conditional text generation task. The reference target sequence $\mathbf{Y}$ is constructed by serializing the scalar quality score $S$, the success label $O$, and the trajectory source label $C$ into a structured textual format. AutoEval-S is trained with supervised fine-tuning by minimizing the negative log-likelihood of the reference tokens:
\begin{equation}
    \mathcal{L}_{\mathrm{SFT}}
    =
    - \sum_{t=1}^{L}
    \log P_\theta
    \left(
    y_t \mid y_{<t}, \mathcal{F}', I_{\mathrm{phys}}
    \right),
    \label{eq:nll_loss}
\end{equation}
where $L$ is the length of the target sequence, $y_t$ is the $t$-th token in $\mathbf{Y}$, and $y_{<t}$ denotes the preceding generated tokens. This formulation enables the evaluator to output quality scores and categorical diagnostic labels in a structured format.

\textbf{AutoEval-P.} AutoEval-P targets CoT diagnostic explanation generation. Starting from the SFT model, we apply \textit{Group Relative Policy Optimization} (GRPO)~\cite{GPRO, guo2025deepseek} to improve the consistency between diagnostic explanations and final structured predictions. Each generated response is evaluated by a hybrid reward that combines score accuracy, task-outcome prediction, auxiliary source-tag prediction, and output-format validity.

For continuous quality scoring, we use a Gaussian soft reward:
\begin{equation}
    R_{\mathrm{score}}
    =
    \exp\left(
    -\frac{(S-\hat{S})^2}{2\sigma^2}
    \right),
    \label{eq:gaussian_reward}
\end{equation}
where $\hat{S}$ is the score parsed from the generated CoT response and $S$ is the reference score. The content reward is
\begin{equation}
    R_{\mathrm{acc}}
    =
    \omega_{\mathrm{score}} R_{\mathrm{score}}
    +
    \omega_{\mathrm{succ}} \mathbb{I}(\hat{O}=O)
    +
    \omega_{\mathrm{src}} \mathbb{I}(\hat{C}=C),
    \label{eq:content_aggregation}
\end{equation}
where $\hat{O}$ and $\hat{C}$ denote the predicted task outcome and auxiliary source tag. A format reward $R_{\mathrm{fmt}}$ is added to encourage parseable responses that follow the required template:
\begin{equation}
    R_{\mathrm{total}}
    =
    (1-\gamma) R_{\mathrm{acc}}
    +
    \gamma R_{\mathrm{fmt}}.
    \label{eq:total_reward}
\end{equation}

For each input $x=(\mathcal{F}, I_{\mathrm{phys}})$, GRPO samples a group of responses from the old policy and normalizes their rewards within the group to compute advantages. The policy is then updated with a KL-regularized objective:
\begin{equation}
\begin{aligned}
\mathcal{J}_{\mathrm{GRPO}}(\theta)
&=
\mathbb{E}_{x\sim\mathcal{D}}
\left[
\frac{1}{G}
\sum_{i=1}^{G}
r_i(\theta) A_i
-
\beta D_{\mathrm{KL}}(x)
\right], \\
r_i(\theta)
&=
\frac{\pi_\theta(y_i|x)}
     {\pi_{\mathrm{old}}(y_i|x)} .
\end{aligned}
\label{eq:optimization_objective}
\end{equation}
Here, $D_{\mathrm{KL}}(x)$ denotes the KL divergence between the current policy and the fixed reference policy. Additional implementation details, including response parsing, group-wise advantage normalization, and the training procedure, are provided in \textbf{Appendix VI}.

\begin{table*}[t]
\centering
\caption{
    \textbf{Comparative Performance Analysis on the Eval-Actions Benchmark.} 
    Results are reported across three protocols: Expert Grading (EG), Rank-Guided (RG), and Chain-of-Thought (CoT). 
    To quantify the domain gap, the zero-shot performance of representative VLMs without Supervised Fine-Tuning (\textit{w/o SFT}) is included. 
    The near-zero correlations (e.g., SRCC $\approx$ 0.02) in these baselines indicate the importance of task-specific fine-tuning.
}
\label{eval_all}
\resizebox{\textwidth}{!}{%
\begin{tabular}{l | c | cc | ccc | ccc} 
\toprule
\multirow{2}{*}{\textbf{Method}} & \multirow{2}{*}{\textbf{Label}} & \multicolumn{2}{c|}{\textbf{Score Prediction}} & \multicolumn{3}{c|}{\textbf{Success Prediction}} & \multicolumn{3}{c}{\textbf{Source Prediction}} \\
\cmidrule(lr){3-4} \cmidrule(lr){5-7} \cmidrule(lr){8-10}
 & & SRCC ($\uparrow$) & $R_{\ell_2}$ ($\downarrow$) & Acc ($\uparrow$) & F1 (\%) ($\uparrow$) & AUC (\%) ($\uparrow$) & Acc (\%) ($\uparrow$) & F1 (\%) ($\uparrow$) & AUC (\%) ($\uparrow$) \\
\cmidrule(lr){1-10}

\multirow{3}{*}{InternVL3.5-4B (w/o SFT )\cite{wang2025internvl3}} 
 & EG  & 0.01 & 31.90 & 56.2 & 66.9 & 51.2 & 46.6 & 51.3 & 50.5 \\
 & RG  & 0.02 & 27.97 & 62.3 & 76.8 & 50.0 & 38.8 & 56.0 & 50.0 \\
 & CoT & -- & -- & -- & -- & -- & -- & -- & -- \\
\midrule
\multirow{3}{*}{QwenVL3-4B (w/o SFT ) \cite{yang2025qwen3}} 
 & EG  & -- & -- & -- & -- & -- & -- & -- & -- \\
 & RG  & -- & -- & -- & -- & -- & -- & -- & -- \\
 & CoT & -- & -- & -- & -- & -- & -- & -- & -- \\
\midrule

\multirow{3}{*}{SmolVLM2.2B \cite{marafioti2025smolvlm}} 
 & EG  & 0.41 & 9.70  & 69.3 & 73.9 & 68.2 & 83.9 & 81.2 & 83.4 \\
 & RG  & 0.39 & 10.57 & 66.4 & 71.3 & 65.5 & 76.5 & 73.1 & 76.1 \\
 & CoT & 0.31 & 12.17 & 62.1 & 67.5 & 61.0 & 68.6 & 66.5 & 68.8 \\
\midrule

\multirow{3}{*}{QwenVL2.5-3B \cite{bai2025qwen2}} 
 & EG  & 0.62 & 7.91  & 76.1 & 81.2 & 73.8 & 98.7 & 98.4 & 98.7 \\
 & RG  & 0.64 & 10.21 & 78.4 & 83.2 & 76.0 & 98.7 & 98.4 & 98.7 \\
 & CoT & 0.46 & 9.25  & 69.1 & 74.3 & 67.6 & 80.6 & 78.1 & 80.4 \\
\midrule

\multirow{3}{*}{InternVL3.5-4B \cite{wang2025internvl3}} 
 & EG  & 0.80 & 3.84 & 90.0 & 92.1 & 88.1 & 94.9 & 93.3 & 94.9 \\
 & RG  & 0.81 & 4.93 & 90.6 & 92.5 & 89.6 & 98.7 & 98.3 & 98.4 \\
 & CoT & 0.63 & 5.68 & 81.7 & 84.2 & 81.0 & 85.0 & 83.6 & 85.0 \\
\midrule

\multirow{3}{*}{QwenVL3-4B \cite{yang2025qwen3}} 
 & EG  & 0.78 & 4.69 & 90.2 & 92.4 & 88.2 & 96.8 & 95.9 & 96.4 \\
 & RG  & 0.82 & 4.55 & 91.0 & 92.9 & 88.8 & 99.1 & 98.9 & 96.4 \\
 & CoT & 0.64 & 5.68 & 81.0 & 84.0 & 80.3 & 85.8 & 83.5 & 85.4 \\
\midrule

\rowcolor{gray!20} \textbf{AutoEval-S (Ours)} & EG  & 0.81 & \textbf{3.45} & 90.6 & 92.8 & 88.5 & 99.1 & 98.7 & 99.0 \\
\rowcolor{gray!20} \textbf{AutoEval-S (Ours)} & RG  & \textbf{0.84} & 3.49 & \textbf{91.0} & \textbf{93.0} & \textbf{90.1} & \textbf{99.6} & \textbf{99.5} & \textbf{99.5} \\
\rowcolor{gray!20} \textbf{AutoEval-P (Ours)} & CoT &  0.70  & 4.45 & 83.0 & 86.4 & 81.2 & 86.9 & 88.7 & 86.2 \\
\bottomrule
\multicolumn{9}{l}{
    \parbox{\linewidth}{
        \footnotesize \itshape 
        $^{\ast}$Note: Dash entries indicate that the zero-shot model did not produce parseable structured outputs under the required evaluation format.
    }
}
\end{tabular} } 
\end{table*}

\section{Experiments and Discussions}
\label{experiment}

AutoEval is evaluated on the densely annotated Eval-Actions Small (EAS) subset. The experiments include two primary evaluation targets and one auxiliary analysis target: \textbf{Fine-Grained Action Quality Assessment} under the EG, RG, and CoT protocols, \textbf{Task Success Detection}, and \textbf{Auxiliary Trajectory-Source Classification} for controlled source-aware analysis of teleoperated and policy-generated executions. All models are evaluated on disjoint EAS splits. Unless otherwise specified, all VLM-based methods use the same task descriptions, RGB frame inputs, and compact kinematic summaries. Depth streams are stored in Eval-Actions as part of the dataset, but are not used as model inputs in the reported experiments.

\subsection{Evaluation Metrics}
\label{sec:metrics}

We evaluate model performance from two aspects: continuous action quality scoring and binary classification. For quality-score prediction under the EG, RG, and CoT protocols, we report Spearman rank correlation coefficient (SRCC) and relative squared error \(R_{\ell_2}\). SRCC measures whether the predicted scores preserve the relative quality ordering of executions and is the primary metric for fine-grained action quality assessment, while \(R_{\ell_2}\) measures the range-normalized absolute prediction error. For task success detection and auxiliary trajectory-source classification, we report Accuracy, F1-score, and AUC. For text-generation VLMs, quality scores and categorical labels are parsed from structured outputs; when class probabilities or token likelihoods are available, they are used to compute AUC. Detailed metric definitions are provided in \textbf{Appendix VII}.

\subsection{Experimental Details}

We compare AutoEval with representative VLM baselines, including SmolVLM2-2.2B~\cite{marafioti2025smolvlm}, QwenVL2.5-3B~\cite{bai2025qwen2}, QwenVL3-4B~\cite{yang2025qwen3}, and InternVL3.5-4B~\cite{wang2025internvl3}, under the EG, RG, and CoT protocols. All standard VLM baselines and AutoEval-S are supervised fine-tuned with LoRA~\cite{hu2022lora} for 20 epochs. Both AutoEval-S and AutoEval-P use QwenVL3-4B as the base backbone; AutoEval-P is initialized from the supervised fine-tuned checkpoint and further optimized with GRPO for CoT-style diagnostic evaluation. The final checkpoint is selected according to validation performance, using \(R_{\ell_2}\) as the primary validation criterion for score prediction. The learning rate is set to \(1\times10^{-4}\), and the batch size is set to 4.

We compare methods under the same VLM visual-input budget. For standard VLM baselines, 8 visual inputs correspond to 8 sampled RGB frames. For AutoEval-S, 8 visual inputs correspond to 8 spatio-temporal mosaics, covering 32 raw frames under the default \(2\times2\) stitching strategy. Thus, AutoEval-S uses the same number of VLM visual inputs as the baselines, while encoding denser local temporal evidence within each input. Full hyperparameter settings and inference costs are provided in \textbf{Appendix VIII}.

\subsection{Experimental Results}

Table~\ref{eval_all} reports the evaluation results under EG, RG, and CoT protocols.

\textbf{Fine-Grained Action Quality Assessment (EG \& RG).}
As shown in Table~\ref{eval_all}, VLMs without SFT achieve near-zero rank correlation with reference labels under both EG and RG, indicating that general-purpose models require task-specific adaptation for execution-quality evaluation. After SFT, all baselines improve significantly. Among them, AutoEval-S achieves the best or tied-best performance under both protocols, reaching SRCC/$R_{\ell_2}$ of \textbf{0.81}/\textbf{3.45} and \textbf{0.84}/\textbf{3.49}, respectively, together with task success detection accuracies of \textbf{90.6\%} and \textbf{91.0\%}. This demonstrates that STA and compact kinematic summaries provide effective representations for execution-quality assessment under a fixed VLM input budget.

Under the RG setting, AutoEval-S achieves \textbf{99.6\%} trajectory source classification accuracy, reflecting strong discriminability of source labels under the controlled EAS split and supporting internal source-aware analysis of trajectories.

\begin{table}[t]
\centering
\caption{Structured generalization on held-out tasks and robot arms.}
\label{tab:structured_generalization}
\small
\setlength{\tabcolsep}{0pt}
\renewcommand{\arraystretch}{1.08}

\begin{tabular*}{\columnwidth}{@{\extracolsep{\fill}}l c c c c c c@{}}
\toprule
\textbf{Split} & \textbf{M.} & \textbf{Lab.} 
& \textbf{SRCC} & $\mathbf{R_{\ell_2}}$ 
& \textbf{Succ.} & \textbf{Src.} \\
\midrule

\multirow{3}{*}{T+A}
& S & EG  &  0.71  & 5.89  & 80.0 & 90.0 \\
& S & RG  & 0.75 & 6.12  & 88.0 & 90.0 \\
& P & CoT & 0.54 & 8.65 & 76.0 & 76.0 \\
\midrule

\multirow{3}{*}{Arm}
& S & EG  & 0.77  & 3.95 & 90.0 & 94.0 \\
& S & RG  & 0.79 & 4.31 & 90.0 & 96.0 \\
& P & CoT & 0.67 & 5.12 & 80.0 & 82.0 \\
\midrule

\multirow{3}{*}{Task}
& S & EG  & 0.75 & 4.11 & 90.0 & 92.0 \\
& S & RG  & 0.76 & 4.74 & 88.0 & 92.0 \\
& P & CoT & 0.66 & 5.89 & 80.0 & 80.0 \\
\midrule

\multirow{3}{*}{RoboMIND-S}
& S & EG  & 0.76 & 4.25 & 90.0 & 92.0 \\
& S & RG  & 0.78 & 4.88 & 90.0 & 94.0 \\
& P & CoT & 0.66 & 6.01 & 82.0 & 82.0 \\

\multirow{3}{*}{Open-X-S}
& S & EG  & 0.81 & 3.59 & 96.0 & 96.0 \\
& S & RG  & 0.83 & 4.13 & 96.0 & 98.0 \\
& P & CoT & 0.73 & 3.77 & 92.0 & 92.0 \\
\bottomrule
\end{tabular*}

\vspace{0.5mm}
\parbox{\columnwidth}{
\footnotesize
\textit{Note:} T+A denotes unseen tasks and unseen arms. Arm and Task denote unseen-arm-only and unseen-task-only splits. S/P denotes AutoEval-S/P. Succ. (Success Prediction) and Src. (Source Prediction) are accuracies (\%).
}
\end{table}

\textbf{CoT Diagnostic Evaluation.}
The CoT protocol evaluates whether the model can generate a diagnostic rationale together with structured predictions. Compared with direct score prediction under EG/RG, this setting imposes stronger constraints because the generated rationale must remain consistent with the visual evidence, quality score, success label, and auxiliary trajectory-source tag. As shown in Table~\ref{eval_all}, finetuned VLM baselines obtain lower score-prediction performance under CoT than under EG or RG; for example, QwenVL3-4B reaches an SRCC of 0.64 under the CoT protocol. AutoEval-P improves the CoT SRCC to \textbf{0.70} and achieves \textbf{83.0\%} task success detection accuracy. This correlation remains lower than that of AutoEval-S under EG/RG protocols, indicating that CoT introduces additional reasoning constraints while still preserving structured evaluability. We therefore treat CoT as a complementary diagnostic signal rather than a replacement for direct score prediction.

\textbf{Summary and Analysis.}
Overall, the results in Table~\ref{eval_all} indicate that task-specific fine-tuning is essential for robotic execution-quality assessment: general VLMs without fine-tuning show near-zero rank correlation with the reference labels, while finetuned models achieve substantial improvements. Based on this, AutoEval-S achieves the best or tied-best performance under both EG and RG protocols, with an SRCC of \textbf{0.84} under RG, demonstrating that Spatio-Temporal Aggregation and compact kinematic summaries provide effective short-term motion information under a fixed VLM input budget. AutoEval-P further improves score prediction under the CoT protocol while providing interpretable diagnostic outputs. Overall, AutoEval provides fine-grained video-based diagnostic signals for execution quality, complementing conventional success-rate evaluation.

\subsection{Structured Generalization Evaluation}
\label{sec:structured_generalization}

To evaluate AutoEval under structured out-of-distribution settings, we construct five 50-trajectory evaluation splits: unseen task and unseen arm (Task+Arm, T+A), unseen arm only (Arm), unseen task only (Task), a cross-dataset split based on RoboMIND (RoboMIND-S), and another cross-dataset split based on Open-X Embodiment (Open-X-S). The first three splits isolate task shift, robot morphology shift, and their joint shift, while RoboMIND-S and Open-X-S provide additional data-distribution shifts.

As shown in Table~\ref{tab:structured_generalization}, AutoEval-S maintains measurable action-quality ranking performance across the first four splits, although its performance decreases compared with the in-distribution EAS test results. In the most challenging T+A setting, where both the task categories and robot arms are excluded from the training split, AutoEval-S obtains SRCC values of \textbf{0.71} and \textbf{0.75} under EG and RG, respectively; under RG, the success and source-label prediction accuracies are \textbf{88.0\%} and \textbf{90.0\%}. When only the arm is held out, AutoEval-S achieves SRCC values of \textbf{0.77}/\textbf{0.79} under EG/RG; when only the task is held out, the corresponding SRCC values are \textbf{0.75}/\textbf{0.76}. These results indicate that AutoEval-S retains relatively stable action-quality ranking ability under single-task or embodiment shifts, while the joint task-and-arm shift introduces a more challenging generalization setting.

On the RoboMIND-S split, AutoEval-S achieves SRCC values of \textbf{0.76} and \textbf{0.78} under EG and RG, with source-label prediction accuracies of \textbf{92.0\%} and \textbf{94.0\%}, suggesting a degree of stability in the sampled cross-dataset setting. In contrast, AutoEval-P under the CoT protocol obtains lower SRCC values of \textbf{0.54--0.67}, indicating that generating diagnostic explanations under distribution shift is more challenging than predicting scalar quality scores. Overall, these structured splits provide preliminary evidence of generalization, while also showing that AutoEval still has room for improvement when task, embodiment, and data distributions shift.

Furthermore, on the Open-X-S split, the evaluation task is simplified because this subset predominantly consists of high-quality, successful trajectories. Under this distribution, AutoEval-S achieves SRCC values of 0.81 and 0.83 under the EG and RG protocols, respectively, with success prediction accuracies of 96.0\% and 96.0\% and source-label prediction accuracies of 96.0\% and 98.0\%. AutoEval-P obtains an SRCC of 0.73 under the CoT protocol, with 92.0\% success and source-label prediction accuracies. Because this subset is dominated by successful trajectories, these results should be interpreted as sampled cross-dataset evidence rather than a comprehensive demonstration of out-of-distribution robustness.

\begin{table}[t]
    \centering
    \caption{Policy-level ranking using AutoEval-S under the EG protocol.
    Each policy--task pair is evaluated with $R=20$ rollout trials. The
    reported score and success rate are averaged over evaluated tasks and
    rollouts.}
    \label{tab:policy_scores}
    \small
    \setlength{\tabcolsep}{4pt}
    \begin{tabular}{lccc}
        \toprule
        \textbf{Policy} & \textbf{Avg. AutoEval-EG} & \textbf{Success Rate} & \textbf{Quality Rank} \\
        \midrule
        $\pi_{0.5}$ (held-out) & \textbf{6.23} & \textbf{72\%} & 1 \\
        $\pi_0$                & 4.47 & 68\% & 2 \\
        ACT                    & 3.70 & 40\% & 3 \\
        RDT                    & 2.43 & 56\% & 4 \\
        DP (held-out)          & 2.12 & 42\% & 5 \\
        \bottomrule
    \end{tabular}

    \vspace{0.5mm}
    \parbox{\columnwidth}{
    \footnotesize
    \textit{Note:} ``Held-out'' denotes policy models whose rollout trajectories are excluded from the AutoEval training split.}
    \vspace{-4mm}
\end{table}

\begin{table*}[t]
\centering
\caption{Ablation on aggregation grid size under EG and RG protocols.}
\label{ablation2}
\begin{tabular}{l | c | cc | ccc | ccc} 
\toprule
\multirow{2}{*}{\textbf{Stitched Frames}} & \multirow{2}{*}{\textbf{Label}} & \multicolumn{2}{c|}{\textbf{Score Prediction}} & \multicolumn{3}{c|}{\textbf{Success Prediction}} & \multicolumn{3}{c}{\textbf{Source Prediction}} \\
\cmidrule(lr){3-4} \cmidrule(lr){5-7} \cmidrule(lr){8-10}
 & & SRCC ($\uparrow$) & $R_{\ell_2}$ ($\downarrow$) & Acc ($\uparrow$) & F1 (\%) ($\uparrow$) & AUC (\%) ($\uparrow$) & Acc (\%) ($\uparrow$) & F1 (\%) ($\uparrow$) & AUC (\%) ($\uparrow$) \\
\cmidrule(lr){1-10}
\rowcolor{gray!20}
 & EG & 0.81 & 3.45 & 90.6 & 92.8 & 88.5 & 99.1 & 98.7 & 99.0 \\
\rowcolor{gray!20}
\multirow{-2}{*}{2 $\times$ 2} & RG & 0.84 & 3.49 & 91.0 & 93.0 & 90.1 & 99.6 & 99.5 & 99.5 \\
\midrule
\multirow{2}{*}{3 $\times$ 3} 
  & EG  & 0.76 & 4.61 & 85.9 & 89.2 & 83.5 & 97.6 & 96.9 & 97.4 \\
  & RG  & 0.77 & 5.57 & 86.7 & 89.7 & 84.7 & 98.9 & 98.5 & 98.7 \\
\midrule
\multirow{2}{*}{4 $\times$ 4} 
  & EG  & 0.60 & 8.28 & 76.4 & 81.6 & 74.0 & 97.3 & 96.5 & 97.1 \\
  & RG  & 0.61 & 13.00 & 74.2 & 81.2 & 69.1 & 97.9 & 97.3 & 97.5 \\
\bottomrule
\end{tabular}
\vspace{-4mm}
\end{table*}

\subsection{Policy-Level Ranking}
\label{sec:policy_ranking}

To illustrate the use of AutoEval for offline policy comparison, we apply AutoEval-S under the EG protocol to rollout videos generated by representative VA/VLA policies, including ACT, $\pi_0$, $\pi_{0.5}$, RDT, and DP, where $\pi_{0.5}$ and DP denote held-out policy families excluded from the AutoEval training split. 
All policies are evaluated on the same 5 tasks with identical task descriptions and comparable initial configurations. For each policy-task pair, we conduct $R=20$ rollouts, and policy-level scores are first averaged within each task and then averaged across tasks.
The AutoEval-EG score reported in Table~\ref{tab:policy_scores} is not an additional human annotation; it is predicted by AutoEval-S trained under the EG protocol. Therefore, this score should be interpreted as a diagnostic quality estimate aligned with the criteria-based expert grading definition in Eval-Actions.

As shown in Table~\ref{tab:policy_scores}, $\pi_{0.5}$ achieves the highest average AutoEval-EG score and the highest success rate, while $\pi_0$ ranks second. Notably, the ranking induced by the EG-style quality score is not identical to the ranking induced by binary success rate. For example, RDT achieves a higher success rate than ACT, but receives a lower average AutoEval-EG score. This discrepancy indicates that policies with higher success rates may still produce low-quality successes involving unstable contacts, repeated retries, inefficient motions, or visible collision-related events. Thus, AutoEval-EG serves as a complementary diagnostic metric to success rate, helping reveal execution-process differences that are not captured by binary task outcomes during offline policy comparison.

\subsection{Ablation Study}
\label{sec:ablation}

We conduct ablation studies to analyze the main design choices of AutoEval, including the grid size of Spatio-Temporal Aggregation, core model components, and input modalities. These results are used to characterize how each component affects diagnostic evaluation performance, rather than treating any single component as a standalone general-purpose temporal modeling solution. More ablation studies can be found in Appendix IX.

\textbf{Aggregation Grid Size.}
We then evaluate the grid size used in the Spatio-Temporal Aggregation strategy of AutoEval-S. As shown in Table~\ref{ablation2}, the $2\times2$ aggregation grid achieves the strongest score-prediction performance among the evaluated grid configurations under both EG and RG, reaching an SRCC of \textbf{0.84} under the RG protocol. Increasing the grid size to $3\times3$ or $4\times4$ includes more frames in each composite image, but reduces the spatial resolution assigned to each frame after resizing to the VLM input size. This trade-off between temporal coverage and spatial detail leads to lower score prediction performance, especially under the $4\times4$ setting, where the SRCC drops to 0.60 and 0.61 under EG and RG, respectively. Therefore, AutoEval-S uses the $2\times2$ configuration as the default setting.

\textbf{Impact of Core Mechanisms.}
Table~\ref{core_mechanisms} evaluates the effects of Spatio-Temporal Aggregation and GRPO under the CoT protocol. Removing Spatio-Temporal Aggregation decreases the SRCC from \textbf{0.70} to 0.67 and increases $R_{\ell_2}$ from \textbf{4.45} to 4.99, indicating that denser temporal evidence remains useful for CoT-style quality assessment, since sparse keyframes may miss short-duration defects such as jitter, hesitation, or repeated retries. Removing GRPO and relying only on supervised fine-tuning leads to an SRCC of 0.68 and an $R_{\ell_2}$ of 5.16. The gain from GRPO is small but consistent in both the correlation and error metrics, suggesting that reward-based optimization can modestly improve the consistency between diagnostic explanations and final scores. The full AutoEval-P achieves the best CoT result in this ablation, with an SRCC of \textbf{0.70} and an $R_{\ell_2}$ of \textbf{4.45}.

\begin{table}[t]
\centering
\caption{Ablation Study on Core Mechanisms under CoT Protocol.
}
\label{core_mechanisms}
\begin{tabular}{l | cc} 
\toprule
\multirow{2}{*}{\textbf{Method}} & \multicolumn{2}{c}{\textbf{Score Prediction}} \\
\cmidrule(lr){2-3}
 & SRCC ($\uparrow$) & $R_{\ell_2}$ ($\downarrow$) \\
\midrule
\textit{w/o} Spatio-Temporal Aggregation & 0.67 & 4.99 \\ 
\textit{w/o} GRPO & 0.68 & 5.16 \\ 
\midrule
\textbf{AutoEval-P} & \textbf{0.70} & \textbf{4.45} \\
\bottomrule
\end{tabular}
\end{table}

\begin{table}[t]
\centering
\caption{Metric-only comparison and AutoEval input ablations under the RG protocol. Metric-only baselines use only explicit physical indicators and robot-state statistics, without RGB frames or VLM reasoning.}
\label{tab:metric_only}
\small
\setlength{\tabcolsep}{4pt}
\begin{tabular}{lcc}
\toprule
\textbf{Evaluator} & SRCC $\uparrow$ & $R_{\ell_2}$ $\downarrow$ \\
\midrule
\multicolumn{3}{l}{\textit{Explicit metric-only baselines}} \\
Metric-MLP  & 0.42 & 10.67 \\
Metric-Rule & 0.36 & 12.18 \\
\midrule
\multicolumn{3}{l}{\textit{AutoEval input ablations}} \\
AutoEval-S ($I_{\mathrm{phys}}$ only) & 0.54 & 11.79 \\
AutoEval-S (RGB only) & 0.81 & 4.16 \\
AutoEval-S (RGB + $I_{\mathrm{phys}}$) & \textbf{0.84} & \textbf{3.49} \\
\bottomrule
\end{tabular}
\end{table}

\textbf{Metric-Only Comparison and Modality Importance.}
We examine whether AutoEval provides information beyond explicit physical metrics. As shown in Table~\ref{tab:metric_only}, metric-only baselines use only rule-defined physical indicators and robot-state statistics (e.g., task completion, normalized execution time, velocity/acceleration variation, and collision-related events), with manually assigned weights and no RGB input or VLM reasoning. Metric-MLP and Metric-Rule achieve SRCCs of 0.42 and 0.36, indicating that explicit physical metrics provide useful but limited cues.

We also evaluate AutoEval-S variants with different input modalities. Removing RGB frames causes a significant performance drop (SRCC 0.84 → 0.54, $R_{\ell_2}$ 3.49 → 11.79), showing that RGB temporal evidence is critical for object interaction, task completion, and collision-related events. Removing the kinematic summary $I_{\mathrm{phys}}$ yields SRCC 0.81, suggesting that substantial quality information can be inferred from visual evidence alone. Including $I_{\mathrm{phys}}$ improves SRCC to 0.84 and reduces $R_{\ell_2}$ to 3.49, indicating that robot-state statistics provide complementary calibration for smoothness and efficiency. Overall, these results show that AutoEval complements explicit physical metrics rather than replacing them: physical metrics provide transparent motion cues, while RGB video supplies task- and object-level semantic evidence.

\section{Conclusion}

VA/VLA robotic manipulation evaluation is formulated as a fine-grained video action-quality assessment (AQA) problem, and the Eval-Actions benchmark and AutoEval framework are introduced. Eval-Actions provides RGB-D recordings, robot-state trajectories, success/failure labels, teleoperated and policy-generated executions, and criteria-based EG, RG, and CoT annotations. These annotations enable diagnostic analysis of execution quality, task success, and trajectory source beyond conventional binary success rates. Built on this benchmark, AutoEval evaluates manipulation executions using RGB temporal evidence and compact kinematic summaries. AutoEval-S preserves short-term motion cues through Spatio-Temporal Aggregation, while AutoEval-P generates CoT-style diagnostic explanations. On the Eval-Actions test split, AutoEval-S achieves SRCC values of \textbf{0.81} (EG) and \textbf{0.84} (RG), and \textbf{99.6\%} trajectory source-label accuracy under RG. This high source-classification accuracy supports source-aware trajectory analysis within the benchmark and suggests that future work may explore source-aware anomaly screening under explicitly controlled protocols. Additional analyses, including criteria-based annotation evaluation, explicit metric-only comparison, structured generalization, and policy-level ranking, indicate that Eval-Actions and AutoEval provide a reproducible video-based diagnostic tool that complements conventional success-rate evaluation and explicit physical metrics.

This work presents several limitations. The current quality scores primarily evaluate execution quality concerning spatial generalization, yet they do not explicitly measure policy-level generalization across varying language instructions, unseen objects, compositional tasks, or long-horizon rollouts. Furthermore, AutoEval currently relies on RGB frames and compact kinematic summaries, meaning the assessment of collision and safety events is not yet fully comprehensive. Future work could integrate RGB-D data, multi-view observations, and force/tactile signals to enhance the diagnostic capabilities for contact-rich tasks. Finally, AutoEval may encounter evaluation failures under severe visual occlusions or during complex long-horizon manipulation tasks.
\bibliographystyle{IEEEtran}
\bibliography{reference}

\end{document}